\documentclass[acmtog,nonacm]{acmart}

\usepackage{booktabs} 

\usepackage[ruled]{algorithm2e} 
\usepackage{multirow}
\usepackage{multicol}

\newcommand{\best}[1]{\textbf{#1}}
\newcommand{\seco}[1]{\underline{#1}}

\SetAlFnt{\small}
\SetAlCapFnt{\small}
\SetAlCapNameFnt{\small}
\SetAlCapHSkip{0pt}

\acmJournal{TOG}

\newcommand{\ie}{i.e.}
\newcommand{\etc}{etc}
\newcommand{\eg}{e.g.}
\newcommand{\etal}{et al.}
\usepackage{threeparttable}
\usepackage{color, colortbl}
\definecolor{Gray}{gray}{0.9}
\begin{document}
\title{Scale-arbitrary Invertible Image Downscaling}

\author{Jinbo Xing}
\affiliation{%
 \institution{The Chinese University of Hong Kong}
 \country{Hong Kong SAR, China}
}
\authornote{Equal contribution}
\author{Wenbo Hu}
\affiliation{%
 \institution{The Chinese University of Hong Kong}
 \country{Hong Kong SAR, China}
}
\authornotemark[1]

\author{Tien-Tsin Wong}
\affiliation{%
 \institution{The Chinese University of Hong Kong}
 \country{Hong Kong SAR, China}
}
\authornote{Corresponding author}
\email{[jbxing,wbhu,ttwong]@cse.cuhk.edu.hk}

\renewcommand\shortauthors{Xing et al.}

\begin{abstract}

Conventional social media platforms usually downscale the HR images to restrict their resolution to a specific size for saving transmission/storage cost, which leads to the super-resolution (SR) being highly ill-posed.
Recent invertible image downscaling methods jointly model the downscaling/upscaling problems and achieve significant improvements.
However, they only consider fixed integer scale factors that cannot downscale HR images with various resolutions to meet the resolution restriction of social media platforms.
In this paper, we propose a {Scale-\textbf{A}rbitrary \textbf{I}nvertible Image \textbf{D}ownscaling \textbf{N}etwork (AIDN)}, to natively downscale HR images with arbitrary scale factors.
%
Meanwhile, the HR information is embedded in the downscaled low-resolution (LR) counterparts in a nearly imperceptible form such that our AIDN can also restore the original HR images solely from the LR images.
The key to supporting arbitrary scale factors is our proposed {Conditional Resampling Module (CRM)} that conditions the downscaling/upscaling kernels and sampling locations on both scale factors and image content.
Extensive experimental results demonstrate that our AIDN achieves top performance for invertible downscaling with both arbitrary integer and non-integer scale factors.
Code will be released upon publication.
\end{abstract}

%
%


%
%



\maketitle
\textbf{Reference Format:}\\
Jinbo Xing, Wenbo Hu, and Tien-Tsin Wong. 2022. Scale-arbitrary Invertible Image Downscaling. \emph{arXiv preprint arXiv:2201.12576.}

\section{Introduction}
\label{sec:intro}

%

\begin{figure}[!t]
  \centering
  \includegraphics[width=1\linewidth]{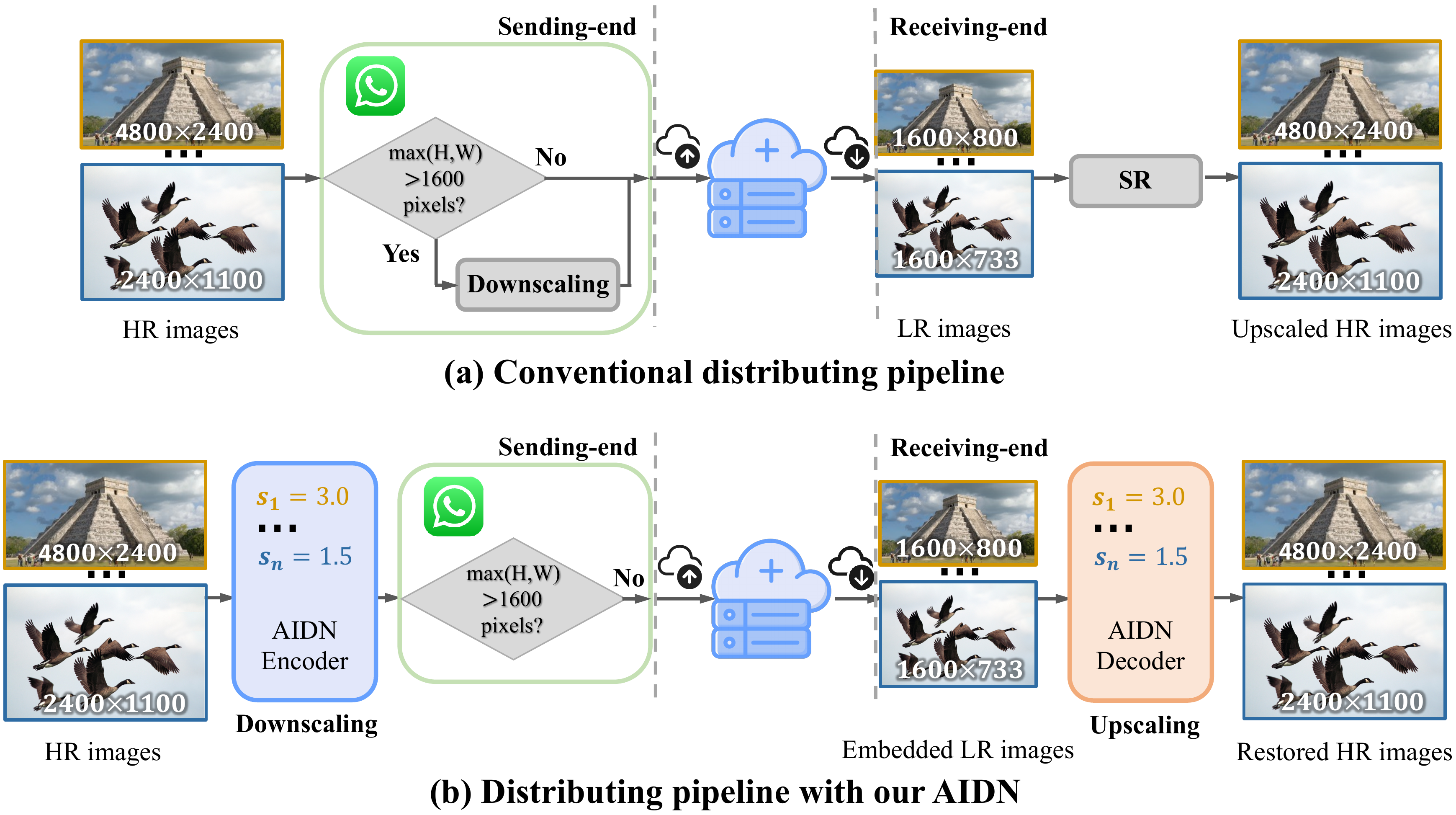}
  \caption{Part (a) shows the conventional pipeline of distributing HR images over the social media platform, \ie, WhatsApp. In the sending-end, the client will downscale HR images to make the height/width of the images equal to or smaller than $1600$ pixels\protect\footnotemark{}~if the resolution of them is too large. And in the receiving-end, image super-resolution (SR) will be employed if users want to explore the details of the distributed images. Part (b) shows the distributing pipeline with our proposed AIDN. In the sending-end, before feeding into the WhatsApp client, the HR images are first downscaled by our AIDN encoder to meet the resolution restriction of the WhatsApp, thus, WhatsApp would not downscale the images anymore. And in the receiving-end, the AIDN decoder can faithfully restore the original HR image whenever necessary.}
  \label{fig:usecase}
\end{figure}

With the rapid development of smartphone cameras, exploding amount of high-resolution (HR) images/videos are produced in our daily life. 
For the transmission/storage cost concern, most social media platforms (\eg, WhatsApp, Messenger, WeChat Moments and etc.) will downscale the HR images to \emph{restrict their resolution to a specific size} when users distribute them over the platform (Figure~\ref{fig:usecase} (a)).
Therefore, image upscaling/super-resolution (SR) is indispensable when receivers want to explore the details of the distributed images.
But the lost information during the downscaling process makes the SR problem highly ill-posed~\cite{dong2015image,glasner2009super,yang2010image}.
To relieve this problem, recent works~\cite{sun2020learned,kim2018task,xiao2020invertible} regard the downscaling/upscaling as a dual problem and jointly optimize them.
By doing so, the information from the HR images can be better preserved or embedded during the downscaling process, which leads to a higher restoration performance for the upscaling counterpart.

\begin{figure}[!t]
  \centering
  \includegraphics[width=0.92\linewidth]{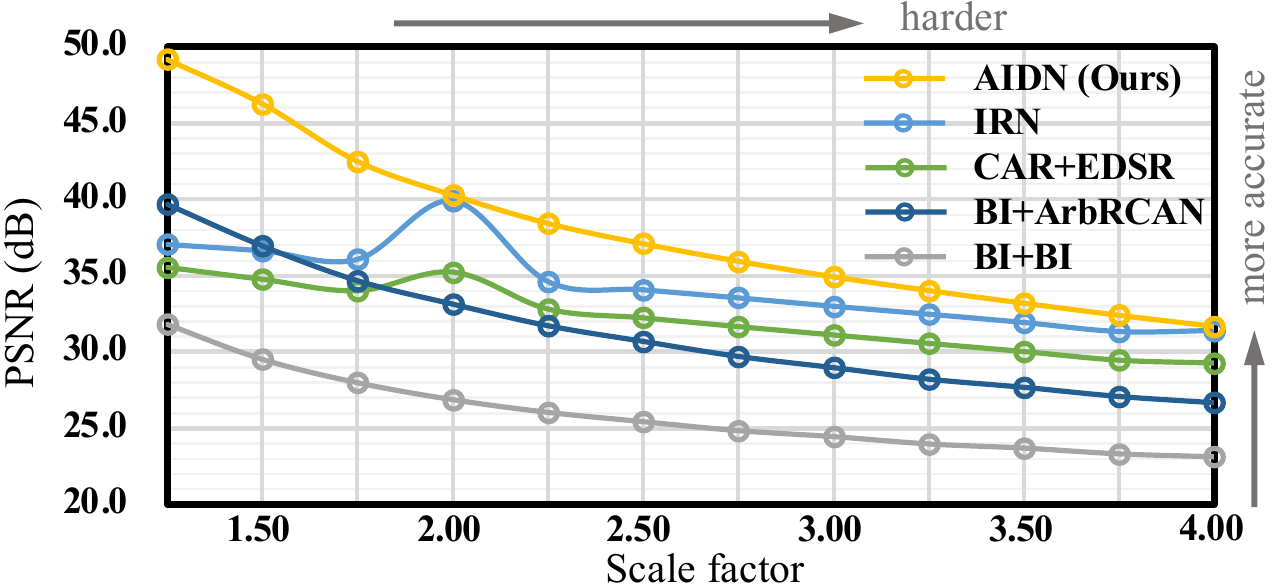}
  \caption{
  Restoration quality \textit{vs.} scale factor for multiple invertible image downscaling solutions, 
  including downscaling/upscaling with both the Bicubic interpolation (BI+BI), state-of-the-art scale-arbitrary SR method (BI+ArbRCAN~\cite{Wang2021Learning}), scale-fixed invertible image downscaling methods (CAR+EDSR~\cite{sun2020learned} and IRN~\cite{xiao2020invertible}), and our proposed \emph{Scale-\textbf{A}rbitrary \textbf{I}nvertible Image \textbf{D}ownscaling \textbf{N}etwork (AIDN)}, on the Urban100~\cite{huang2015single} dataset.
  }
  \label{fig:intro}
\end{figure}

However, existing works only consider a fixed integer scale factor, \eg, $\times$2, $\times$3, or $\times$4.
%
In real-world scenarios, the scale factor should be arbitrary, since the social media platform will downscale images with various resolutions to meet the resolution restriction (Figure~\ref{fig:usecase} (a)). 
Although with certain pre-/post-processing operations (including the Bicubic interpolation~\cite{mitchell1988reconstruction}) on the input/output images from \emph{multiple} trained scale-fixed invertible models~\cite{xiao2020invertible,sun2020learned}, one can indirectly and virtually achieve the arbitrary scale factor, this also means numerous trained models have to be stored in place. 
Moreover, the experiment shows such an indirect solution cannot produce satisfactory results (Figure~\ref{fig:intro}), and its performance fluctuates across the scale factors.

\footnotetext{The resolution restriction of WhatsApp is determined by experiments, and other social media platforms (\eg, Messenger, WeChat Moments and etc.) share similar rules.}

In this paper, we propose a \emph{universal} encoder-decoder styled \emph{Scale-\textbf{A}rbitrary \textbf{I}nvertible Image \textbf{D}ownscaling \textbf{N}etwork (AIDN)}, to downscale/upscale images with arbitrary scale factors. 
%
The encoder network is designed for the downscaling problem while the decoder network is for the upscaling task, and the encoder-decoder is trained jointly to address their duality.
As shown in Figure~\ref{fig:usecase} (b), in the sending-end, the encoder of our AIDN can natively downscale HR images using arbitrary required scale factors to meet the resolution restriction of a particular social media platform, and embed the HR information to the LR image in a nearly imperceptible form;
while in the receiving-end, the decoder of our AIDN can consume the embedded information to faithfully restore the original HR image, solely from the LR counterpart, whenever receivers want to explore the details of the distributed images.
The key to supporting arbitrary scale factors is our proposed \emph{Conditional Resampling Module (CRM)}.
It can dynamically resample the feature map to target resolution by parameterizing the downsampling/upsampling kernels and sampling locations conditioned on both the target scale factor and the image content.
The content-adaptive characteristic allows the resampling operation to adapt to the textural/structural content at both training and inference time, so that we can produce visually pleasant results.
Our proposed CRM can be easily applied to multiple existing backbones in the encoder-decoder network to enable the scale-arbitrary invertible image downscaling.

We evaluated our AIDN on multiple public datasets, both quantitatively and qualitatively. 
Experiment results show our AIDN achieves top performance for both integer (\eg, $\times$2, $\times$3 and $\times$4) and arbitrary non-integer scale factors (\eg, $\times$1.60, $\times$2.75 and $\times$3.20).
Moreover, the performance changes smoothly (\ie~more predictable), without fluctuation, across the scale factors, as shown in Figure~\ref{fig:intro}. 
Also, the visualization of routing weights confirms our CRM can dynamically adjust resampling kernels for various scale factors and different image content (Section~\ref{sec:routing}).
Our contributions are summarized below.
\begin{itemize}

    \item To the best of our knowledge, this is the first attempt to tackle the \emph{scale-arbitrary invertible image downscaling} problem with a single encoder-decoder network.

    \item We propose a \emph{Conditional Resampling Module (CRM)} to dynamically resample feature maps to the target resolution according to both the required scale factor and the image content. It can be easily applied to existing SR backbones to achieve scale-arbitrary invertible image downscaling.

    \item Extensive experiments demonstrate our AIDN achieves top performance for invertible downscaling with both arbitrary integer and non-integer scale factors.
    Moreover, the amount of parameters of our AIDN is significantly reduced ($\sim$1/10) compared to conventional SR networks, which indicates the high efficiency of our simple yet effective system design. 
\end{itemize}


\section{Related Work}
\subsection{Image Rescaling}
%
Image downscaling and upscaling/super-resolution (SR) have been studied with a long history.
Downscaling aims at reducing the resolution of images.
%
It can be performed by resampling together with interpolation, \eg, the Bilinear and Bicubic~\cite{mitchell1988reconstruction} interpolation, which enjoys high efficiency but usually incurs visual artifacts, such as aliasing, ringing, blurring, \etc.
To tackle these problems, several detail-preserved and perceptual-quality-oriented approaches~\cite{kopf2013content,oeztireli2015perceptually,liu2017l_,weber2016rapid} are proposed.
On the other hand, SR tries to restore the high-resolution (HR) image from its low-resolution (LR) counterpart.
Thanks to deep learning, we have witnessed significant progress~\cite{jo2021practical,haris2018deep,dong2014learning,kim2016accurate,zhang2018residual,zhang2018image,dai2019second,niu2020single,Wang2021Learning,zhong2018joint,liang2021swinir} on the SR problem. 
%
%
Recently, several works propose to learn a single network for scale-arbitrary SR, using meta-learning~\cite{hu2019meta}, local implicit image function~\cite{chen2021learning}, and scale-aware upsampling~\cite{Wang2021Learning}.

%
The downscaling and the upscaling act as a dual problem in nature, however, all the above works independently model them.
Thus, they may be sub-optimal when working together, which is a common scenario when distributing HR images over social media platforms.  
Differently, we model the downscaling and upscaling problems as a universal encoder-decoder network and optimize them jointly.
By doing so, the downscaling and the upscaling can mutually reinforce each other to produce more visually satisfying results for both downscaled and upscaled images.

\begin{figure*}[!ht]
  \centering
   \includegraphics[width=1.0\linewidth]{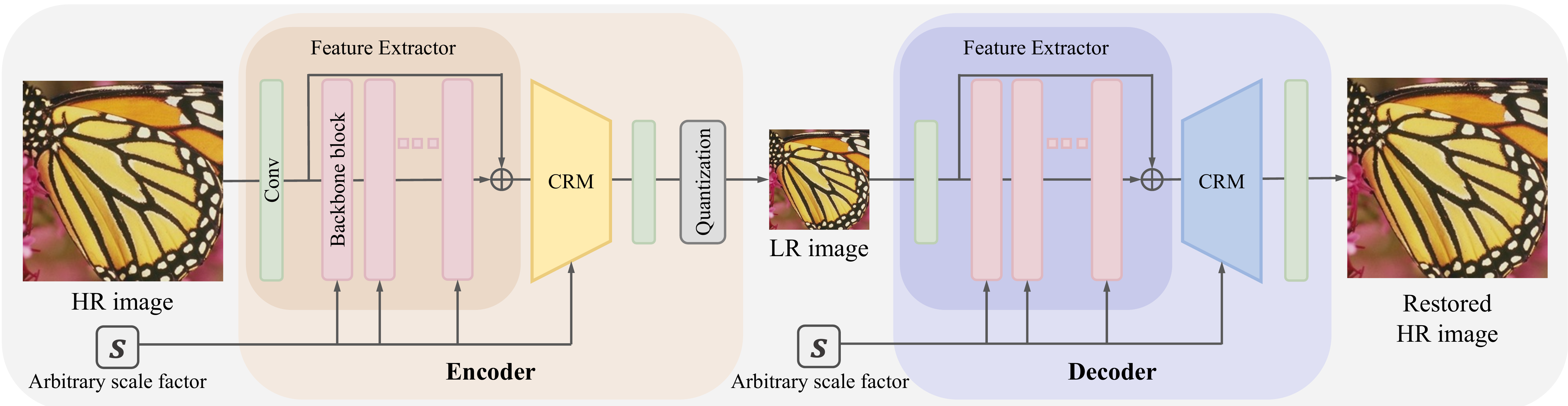}
	\vspace*{-6mm}
   \caption{Overview of the \emph{Scale-\textbf{A}rbitrary \textbf{I}nvertible Image \textbf{D}ownscaling \textbf{N}etwork (AIDN)}.
   Given an HR image to be distributed and the required arbitrary scale factor $s$, the encoder can downscale the HR image to an LR image to meet the resolution restriction of social media platform; meanwhile, the decoder network can restore the original HR image solely from the LR counterpart, whenever users want to explore the details of the distributed image. The CRM is our proposed conditional resampling module to resample feature maps with arbitrary scale factors. 
   }
   \vspace*{-3mm}
   \label{fig:overview}
\end{figure*}

\subsection{Invertible Image Conversion}
The goal of invertible image conversion is to build an invertible transformation between certain visual content and an embedding image, where the original content can be restored from the embedding image whenever necessary~\cite{cheng2021iicnet}.
As a pioneer, Xia \etal~\cite{xia2018invertible} propose an auto-encoder-style network to encode the color information into the generated grayscale image, from which the original color image can be decoded back.
Furthermore, several works build the invertible transformation between color and halftone images~\cite{xia-2021-inverthalf}, binocular and monocular videos~\cite{hu2020mononizing}, short videos and key-frame~\cite{zhu2020video}, multiview images and JPEG image~\cite{wu2021embedding}, camera raw data and sRGB image~\cite{xing2021invertible}.
And recently, Cheng \etal~\cite{cheng2021iicnet} present a generic invertible neural network (INN)~\cite{dinh2014nice,dinh2016density,kingma2018glow,behrmann2019invertible} based framework for multiple invertible image conversion problems.

Image downscaling/upscaling can also be formulated as invertible image conversion.
Kim \etal~\cite{kim2018task} present a task-aware image downscaling (TAD) method to jointly optimize the downscaling and upscaling networks as a united task.
Li \etal~\cite{li2018learning} propose to use a CNN to estimate compact-resolution images (CNN-CR), and then leverage a specified or learned SR method to restore the HR images;
Recently, Sun \etal~\cite{sun2020learned} present a learnable image downscaling method based on content-adaptive resampler (CAR) that can be jointly trained with existing SR networks, and IRN~\cite{xiao2020invertible} adopts the INN to model the invertible image rescaling task as a bijective mapping from HR image to LR image while capturing the distribution of lost information using a latent variable.
Although these works demonstrate the effectiveness of invertible image downscaling, they can only downscale/upscale images with fixed integer scale factors, \eg, $\times$2 and $\times$4. 
Supporting arbitrary scale factors is crucial in real-world scenarios, as shown in Figure~\ref{fig:usecase}.
%
Different from the above methods, our method can downscale images with arbitrary scale factors and faithfully recover the HR images.

\vspace{-2mm}
\section{Methodology}
\subsection{Overview}
Overall, as shown in Figure~\ref{fig:overview}, our \emph{Scale-\textbf{A}rbitrary \textbf{I}nvertible Image \textbf{D}ownscaling \textbf{N}etwork (AIDN)} contains an encoder $E_{\theta}$ and a decoder $D_{\phi}$, where $\theta$ and $\phi$ denote their parameters, respectively.
Given a high-resolution (HR) image to be shared through social media platforms, $I_\text{HR}$, and the required arbitrary scale factor to meet the resolution restriction of social media platforms, $s \in (1,4]$, the encoder network $E_{\theta}$ can downscale the $I_\text{HR}$ to produce a low-resolution (LR) image, $\hat{I}_\text{LR}$, which looks like the reference LR image $I_\text{LR}$ (say, the bicubic-downsampled~\cite{mitchell1988reconstruction} image, without loss of generality).
The produced LR image $\hat{I}_\text{LR}$ has the same 8-bit precision as conventional images for compatibility with current platforms.
And importantly, the decoder network $D_{\phi}$ can accurately restore the HR image $\hat{I}_\text{HR}$ solely from the LR image $\hat{I}_\text{LR}$, whenever users want to explore the details of the distributed image.
%

We design a universal encoder-decoder network for downscaling/upscaling with arbitrary scale factors.
%
To leverage the duality of the downscaling and the upscaling, we train the encoder and decoder jointly with the supervision on both $\hat{I}_\text{LR}$ and $\hat{I}_\text{HR}$.
By doing so, the information from HR images can be embedded in the $\hat{I}_\text{LR}$ and the decoder network can consume it to restore the high-quality $\hat{I}_\text{HR}$.


\vspace{-3mm}
\subsection{Network Architecture}
As shown in Figure~\ref{fig:overview}, the encoder and decoder sub-networks in our AIDN share a similar structure, where the encoder network contains the feature extractor, the \emph{Conditional Resampling Module} (CRM), and the quantization layer; and decoder network consists of the feature extractor and the CRM.

\vspace{-2.5mm}
\paragraph{Feature extractor.}
Considering the duality of the downscaling and the upscaling, we adopt the same architecture for feature extractors in the encoder $E_{\theta}$ and decoder $D_{\phi}$ networks.
As EDSR~\cite{lim2017enhanced} shows powerful capability for feature extraction in the SR task, we adopt a similar network structure for the feature extractor.
For the efficiency concern, we employ the EDSR-baseline structure as the backbone block, thus, the amount of parameters of our whole network is only 3.8M, as shown in Table~\ref{tab:SOTA_comparison}. 
It consists of a convolutional layer as the head and a series of residual blocks (Conv+ReLU+Conv)~\cite{he2016deep} as the backbone structure to learn the residual components.
Note that many other types of backbone blocks are also applicable to our framework, \eg, RDN~\cite{zhang2018residual} and RCAN~\cite{zhang2018image}.
To better extract scale-adaptive features for our goal, invertible image downscaling with arbitrary scale factors, we also equip the backbone block with the scale-aware feature adaption module~\cite{Wang2021Learning} that takes the feature map from the previous layer and the required arbitrary scale factor as input.
More details about the feature extractor can be found in the \emph{Supplementary Material}.

\vspace{-2.5mm}
\paragraph{Conditional resampling module (CRM)}
After extracting features from $I_\text{HR}$ or $\hat{I}_\text{LR}$, we need to resample the feature map to the target resolution with arbitrary scale factors, $s \in (1.0,4.0]$.
Previous invertible image downscaling methods~\cite{sun2020learned,kim2018task,xiao2020invertible} adopt either PixelShuffle~\cite{shi2016real} or Haar transformation as the resampling module, which, however, inherently only serves a fixed integer scale factor $r$, \ie, $r \in \{2, 3, 4, \; \dots\}$ for the PixelShuffle and $r \in \{2, 4, 8, \; \dots\}$ for the Haar transformation.

Inspired by the conditional convolution~\cite{tian2020conditional,zhang2020dynet,yang2019condconv,chen2020dynamic} that conditions convolutional kernels on instances for visual recognition, we propose a Conditional Resampling Module (CRM) to dynamically downsample/upsample feature maps to the target resolution by parameterizing the resampling kernels and sampling locations conditioned on both the required scale factor and the image content.
Besides, Wang~\etal~\cite{Wang2021Learning} also presents a scale-aware upsampling layer used for SR networks to support arbitrary scale factors.
%
%
Unlike the scale-aware upsampling layer, our CRM can be used for both downscaling and upscaling, and is not only scale-aware but also content-adaptive.
The content-adaptive characteristic allows our method better deal with various textural/structural patterns at both training and inference time, so that our method can produce visually satisfying results, as to be demonstrated in Sec.~\ref{sec:exp}.

\begin{figure}[!t]
	\centering
	\includegraphics[width=0.99\linewidth]{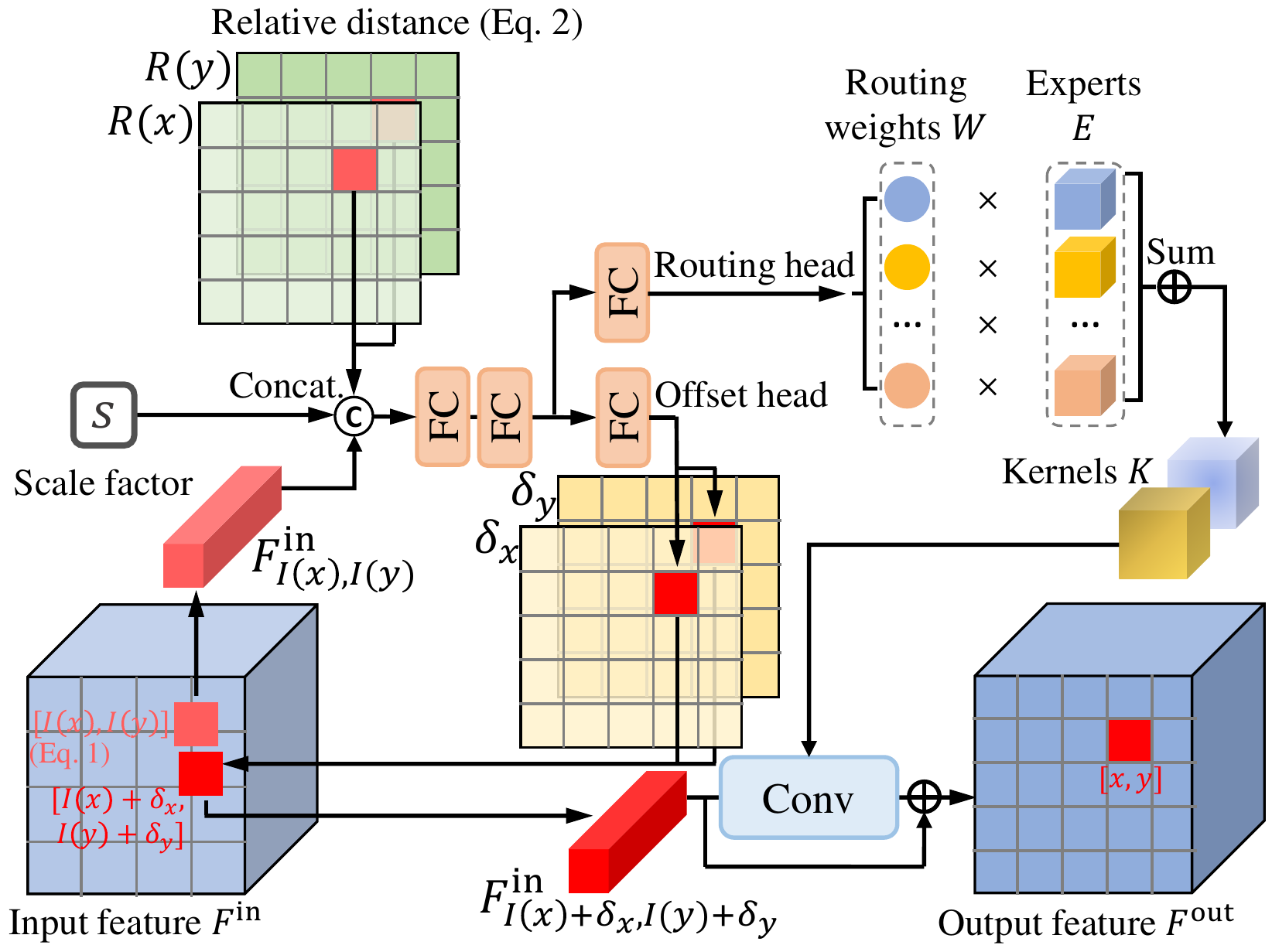}

	\caption{Conditional Resampling Module (CRM). Given the input feature $F^\text{in}$ and required arbitrary scale factor $s$, our CRM can dynamically resample $F^\text{in}$ according to scale factor and image content, for producing the output feature $F^\text{out}$.}

	\label{fig:CRM}
	  \vspace*{-4mm}
\end{figure}

More concretely, as shown in Figure~\ref{fig:CRM}, given the input feature map $F^\text{in}$ and the required arbitrary scale factor $s$, the goal of Conditional Resampling Module (CRM) is to produce the feature map $F^\text{out}$ with the required resolution.
As CRM works for downscaling and upscaling similarly, we only explain the upscaling procedure below.
Taking the computation of $F^\text{out}$ at coordinate $[x,y]$, $F^\text{out}_{x,y}$, as an example, we first project $[x,y]$ to the coordinate of $F^\text{in}$, $\left[I(x),I(y)\right]$, as:
\begin{equation}
	I(\sigma) = \frac{\sigma+0.5}{s}-0.5,
	\quad \sigma \in \{x,y\}.
\end{equation}
Then, we can query a feature vector of $F^\text{in}$ at $[I(x),I(y)]$, $F^\text{in}_{I(x),I(y)}$.
The feature querying can be achieved by interpolation, where we adopt the bilinear interpolation in our implementation.
For each projected coordinate, we also compute a relative distance vector $\left[R(x),R(y)\right]$,
\begin{equation}
 	R(\sigma) = I(\sigma) - \text{floor}\left(\frac{\sigma+0.5}{s}\right),
 	\quad \sigma \in \{x,y\}.
\end{equation}
Next, $s$, $F^\text{in}_{I(x),I(y)}$, and $[R(x),R(y)]$ are concatenated together and fed to two Fully-Connected (FC) layers for feature extraction and aggregation.
The aggregated feature is then passed to two heads: the offset head for predicting the final sampling location offset $[\delta_x,\delta_y]$; and the routing head to predict the routing weights $W$ for the experts $E$.
After that, we query a new feature vector $F^\text{in}_{I(x)+\delta_x,\; I(y)+\delta_y}$ using the predicted offset $[\delta_x,\delta_y]$; and use the predicted routing weights ${W}$ to combine the experts ${E}$, which are learnable parameters, for producing the final resampling kernels, ${K} = {W} \cdot {E}$.
Finally, we can compute the required result $F^\text{out}_{x,y}$ as:
\begin{equation}
	F^\text{out}_{x,y} = {K} \ast F^\text{in}_{I(x)+\delta_x,\; I(y)+\delta_y} +
	F^\text{in}_{I(x)+\delta_x,\; I(y)+\delta_y},
\end{equation}
where $\ast$ is the convolution operator.
We can find that the kernels and sampling locations are both conditioned on the scale factor and the image content; thus, the produced output feature can be adaptive to both scale and content.

\vspace{-1mm}
\paragraph{Quantization layer.} 
The pixel value output by the convolutional layer is inherently a continuous floating-point number.
But we require the downscaled image $\hat{I}_{\text{LR}}$ to be compatible with current platforms, which should be in an 8-bit format, \ie, integers in the range of [0, 255].
It means we have to quantize the floating-point network-output values to 8-bit integers for producing 
$\hat{I}_{\text{LR}}$.
Such operation is unfortunately not differentiable that hinders the end-to-end training of the encoder-decoder.
To this end, several techniques~\cite{balle2016end,balle2016end_2,bengio2013estimating,theis2017lossy} were proposed in the image compression field.
%
Similarly to~\cite{nakanishi2018neural}, we approximate the non-differentiable round operation with a soft version:
\begin{equation}
    \text{round}_{\text{soft}}(x) = x - \alpha \frac{\sin(2\pi x)}{2\pi},
    \label{eqn:roundsoft}
\end{equation}
where $\alpha$ is set to $0.5$ in our experiments.
The quantization layer adopts the conventional \textit{round()} function and the gradient of Equation~\ref{eqn:roundsoft} in the forward and backward passes, respectively. 
\begin{table*}[!t]
  \centering
    \caption{
  Quantitative results (PSNR) of the reconstructed HR images produced by multiple methods with various scale factors on the five benchmark datasets.
  Methods marked with *, $\dag$, and $\ddag$ are the type I, II, and III methods explained in Sec.~\ref{subsec:HR}, respectively.
  The \best{best} and \seco{second-best} results are marked in bold and underline, respectively.  
  }

    \label{tab:SOTA_comparison}
  \begin{threeparttable}
  \setlength{\tabcolsep}{3.5pt}
  \resizebox{\textwidth}{!}{
  \begin{tabular}{l|c|cc|cc|cc|cc|cc}
  \toprule[1pt]
  
  \multicolumn{1}{c|}{Method} & Param. & \multicolumn{2}{c|}{Set5} & \multicolumn{2}{c|}{Set14} & \multicolumn{2}{c|}{B100} & \multicolumn{2}{c|}{Urban100} & \multicolumn{2}{c}{DIV2K} \\
  \multicolumn{1}{c|}{Downscaling+Upscaling}            & (M) & $\times$2    &  $\times$1.6 & $\times$2    & $\times$1.65 & $\times$2    & $\times$1.4  & $\times$2    & $\times$1.95 & $\times$2    & $\times$1.7  \\
  \hline
  Bicubic + Bicubic                                     &  -  & 33.66 & 36.10 & 30.24 & 31.83 & 29.56 & 32.95 & 26.88 & 27.05 & 31.01 & 32.46 \\
  
  Bicubic + EDSR-$\times$2\cite{lim2017enhanced}\tnote{*}& 40.7& 38.19 & 40.39 & 33.95 & 35.95 & 32.36 & 36.79 & 32.95 & 32.69 & 35.03     & 36.95      \\
  
  Bicubic + ArbEDSR\cite{Wang2021Learning}\tnote{$\dag$}& 39.2& 38.19 & 40.64 & 34.05 & 36.22 & 32.37 & 36.92 & 33.02 & 33.30 & -     &-       \\
  
  
  TAD + TAU(-$\times$2)\cite{kim2018task}\tnote{$\ddag$}            &  -  & 38.46 & -     & 35.52 & -     & 36.68 & -     & 35.03 & -     & 39.01 & -     \\
  
  CNN-CR + CNN-SR(-$\times$2)\cite{li2018learning}\tnote{$\ddag$}   &  -  & 38.88 & -     & 35.40 & -     & 33.92 & -     & 33.68 & -     & -     & -     \\
  
  CAR + EDSR(-$\times$2)\cite{sun2020learned}\tnote{$\ddag$}        & 51.1& 38.94 & 40.09 & 35.61 & 36.45 & 33.83 & 36.41 & 35.24 & 33.28 & 38.26 & 34.29 \\
  
  IRN-$\times$2\cite{xiao2020invertible}\tnote{$\ddag$}           & \best{1.7} & \seco{43.99} & \seco{43.42} & \seco{40.79} & \seco{39.24} & \best{41.32} & \seco{39.63} & \seco{39.92} & \seco{35.28} & \best{44.32} & \seco{42.00} \\
  \rowcolor{Gray}
  AIDN (Ours)                                           & \seco{3.8} & \best{44.13} & \best{48.81} & \best{40.81} & \best{44.25} & \seco{40.72} & \best{52.11} & \best{40.28} & \best{39.27} &\seco{44.12}     & \best{47.49}     \\
  
  \midrule[0.8pt]
    &  &                                                           $\times$3 & $\times$2.75 & $\times$3    & $\times$2.8  & $\times$3    & $\times$2.2  & $\times$3    & $\times$2.35 & $\times$3    & $\times$2.55 \\
  \hline
  Bicubic + Bicubic                                     &  -  & 30.39 & 31.06 & 27.55 & 27.84 & 27.21 & 28.88 & 24.46 & 25.72 & 28.22 & 29.27  \\
  
  Bicubic + EDSR-$\times$3\cite{lim2017enhanced}\tnote{*}& 40.7& 34.68 & 35.35 & 30.53 & 30.90 & 29.27 & 31.38 & 28.82 & 30.91 & 31.26     &32.69  \\
  
  Bicubic + ArbEDSR\cite{Wang2021Learning}\tnote{$\dag$}& 39.2& 34.73 & 35.34 & 30.61 & 31.04 & 29.30 & 31.46 & 28.90 & 31.11 & -     &-  \\
  
  
  CNN-CR + CNN-SR(-$\times$3)\cite{li2018learning}\tnote{$\ddag$}   &  -  & 35.13 & -     & 31.33 & -     & 30.26 & -     & 28.81 & -     & -     & -     \\
  
  CAR + EDSR(-$\times$4)\cite{sun2020learned}\tnote{$\ddag$}        & 51.1& 36.13 & 36.69 & 32.52 & 33.04 & 31.29 & 33.56 & 31.12 & 32.59 & 34.15 & 35.84     \\
  
  IRN-$\times$4\cite{xiao2020invertible}\tnote{$\ddag$}           & \seco{4.4} & \seco{38.41} & \seco{39.10} & \seco{35.02} & \seco{35.60} & \seco{34.03} & \seco{36.79} & \seco{33.00} & \seco{34.41} & \seco{37.43} & \seco{38.84}     \\
  \rowcolor{Gray}
  AIDN (Ours)                                           & \best{3.8} & \best{38.70} & \best{39.64} & \best{35.52} & \best{36.23} & \best{34.32} & \best{38.68} & \best{34.93} & \best{37.86}     & \best{37.96}     & \best{40.03}     \\
  
  \midrule[0.8pt]
    &  &                                                           $\times$4 & $\times$3.1  & $\times$4    & $\times$3.2  & $\times$4    & $\times$3.55  & $\times$4    & $\times$3.7 & $\times$4    & $\times$3.65 \\
  \hline
  Bicubic + Bicubic                                     &  -  & 28.42 & 29.89 & 26.00 & 26.98 & 25.96 & 26.32 & 23.14 & 23.38 & 26.66 & 27.10 \\
  
  Bicubic + EDSR-$\times$4\cite{lim2017enhanced}\tnote{*}& 40.7& 32.47 & 34.25 & 28.81 & 29.95 & 27.73 & 28.25 & 26.65 & 27.06 & 29.25 & 29.92     \\
  
  Bicubic + ArbEDSR\cite{Wang2021Learning}\tnote{$\dag$}& 39.2& 32.51 & 34.48 & 28.83 & 30.07 & 27.74 & 28.30 & 26.62 & 27.12 & -     & -     \\
  
  
  TAD + TAU(-$\times$4)\cite{kim2018task}\tnote{$\ddag$}            &  -  & 31.81 & -     & 28.63 & -     & 28.51 & -     & 26.63 & -     & 31.16 & -     \\
  
  CAR + EDSR(-$\times$4)\cite{sun2020learned}\tnote{$\ddag$}        & 51.1& 33.88 & 35.96 & 30.31 & 32.06 & 29.15 & 30.18 & 29.28 & 29.59 & 32.82 & 33.20     \\
  
  IRN-$\times$4\cite{xiao2020invertible}\tnote{$\ddag$}           & \seco{4.4} & \best{36.19} & \seco{38.23} & \best{32.67} & \seco{34.50} & \best{31.64} & \seco{32.56} & \seco{31.41} & \seco{31.48} & \best{35.07} & \seco{35.71}     \\
  \rowcolor{Gray}
  AIDN (Ours)                                           & \best{3.8} & \seco{36.06} & \best{38.38} & \seco{32.57} & \best{34.85} & \seco{31.50} & \best{32.58} & \best{31.68} & \best{32.57} & \seco{34.94}     & \best{35.85}     \\
  
  \bottomrule[1pt]
  \end{tabular}}

  \end{threeparttable}

\end{table*}


\subsection{Loss Function}
Following Xiao~\etal~\cite{xiao2020invertible}, we adopt two loss terms to drive the network training: 
\begin{align}
	\mathcal{L}_{G}(\theta) &= \mathbb{E}_{I_{\text{HR}}\in\mathcal{I}}\left\{\|f(I_{\text{HR}})-\hat{I}_\text{LR}\|_2^2\right\}   \\
	\mathcal{L}_{I}(\theta, \phi) &= \mathbb{E}_{I_{\text{HR}}\in\mathcal{I}}\left\{\|I_\text{HR}-\hat{I}_\text{HR}\|_1\right\},
\end{align}
where the \emph{guidance loss} $\mathcal{L}_{G}(\theta)$ is to supervise the produced downscaled image $\hat{I}_\text{LR}$ to be similar with the conventional LR image that is generated by the Bicubic interpolation $f(\cdot)$ from the original HR image $I_\text{HR}$; and the \emph{invertibility loss} $\mathcal{L}_{I}(\theta, \phi)$ is to constrain the reconstructed HR image $\hat{I}_\text{HR}$ to match $I_\text{HR}$. Here $\mathbb{E}$ denotes the average operator over all images in training dataset $\mathcal{I}$; $\theta$ and $\phi$ denote the parameters of encoder and decoder, respectively.
Note that $\mathcal{L}_{I}(\theta, \phi)$ effectively imposes constraints over the parameters of both the encoder and decoder, since they are jointly trained.
Altogether, we optimize the proposed AIDN by minimizing the \emph{total loss} $\mathcal{L}(\theta, \phi)$, 
\begin{equation}
	\mathcal{L}(\theta, \phi) = \lambda\mathcal{L}_{G}(\theta) + \mathcal{L}_{I}(\theta, \phi),
\end{equation}
where $\lambda$ is a weight for balancing the two terms and set to be $1.0$ in our experiments.

\section{Experiments}
\label{sec:exp}

\subsection{Dataset and Settings}
We employed the DIV2K dataset~\cite{agustsson2017ntire} to train our model, which contains 900 high-quality 2K resolution images.
We followed the official training and validation set splits.
Besides, we evaluated our AIDN on additional four benchmark datasets, \ie, Set5~\cite{bevilacqua2012low}, Set14~\cite{zeyde2010single}, B100~\cite{martin2001database} and Urban100~\cite{huang2015single}. 
%
%

During training, we adopted scale factors varying from $1.0$ to $4.0$ with a stride of $0.1$, \ie, $\mathcal{S}=\{1.1,1.2,...,4\}$.
To address the varying difficulties of different scale factors, we randomly sampled $s$ with probability of $\frac{s^2}{\sum_{\mathcal{S}}s^2}$ from $\mathcal{S}$.
We leave other training details in the \emph{Supplementary Material}, due to the space constraint.



\begin{figure*}[!t]
	\centering
	\includegraphics[width=1\linewidth]{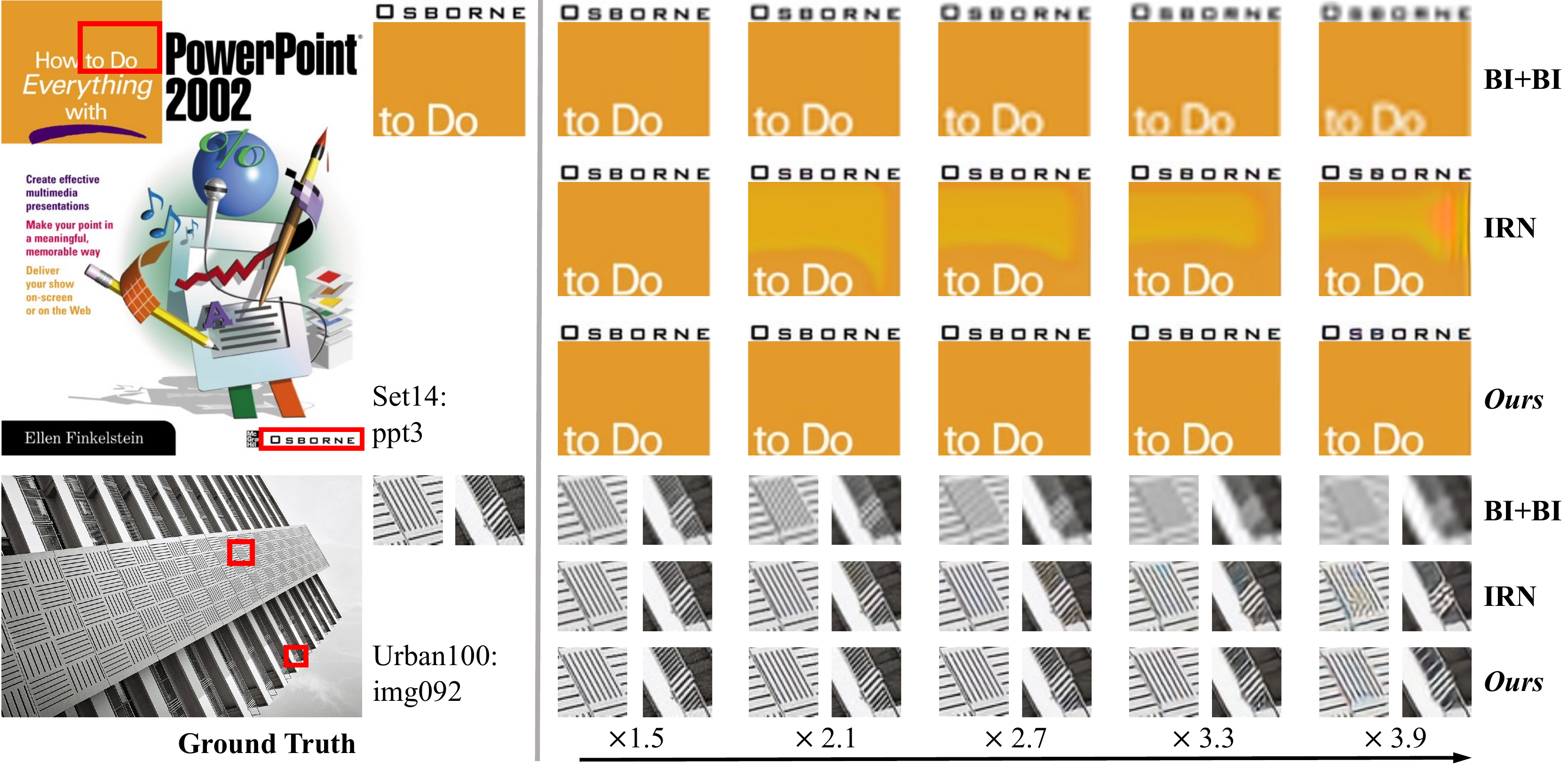}
\vspace{-8mm}
	\caption{Visual comparisons of the reconstructed HR images produced by the Bicubic interpolation (BI+BI), IRN~\cite{xiao2020invertible}, and our AIDN with various non-integer scale factors (from $\times$1.5 to $\times$3.9). The test images are sampled from the Set14~\cite{zeyde2010single} and Urban100~\cite{huang2015single} datasets.
	}
\vspace{-4mm}
	\label{fig:visual}
\end{figure*}

\vspace{-1.5mm}
\subsection{Evaluation on Reconstructed HR Images}
\label{subsec:HR}
First, we evaluated the quality of the reconstructed HR images $\hat{I}_\text{HR}$ for our AIDN and compared it with three types of state-of-the-art solutions, including:

\begin{itemize}\sloppy

  \item[I.] downscaling by the Bicubic interpolation and upscaling by the scale-fixed SR methods, \ie, Bicubic+EDSR-$\times$2/$\times$3/$\times$4~\cite{lim2017enhanced}, and further bicubic-downscaling the output to the target resolution with required non-integer scale factors;

  \item[II.] downscaling by the Bicubic interpolation and upscaling by the scale-arbitrary SR method, \ie, Bicubic+ArbEDSR~\cite{Wang2021Learning}; and

  \item[III.] scaled-fixed invertible image downscaling models, \ie, TAD+TAU~\cite{kim2018task}, CNN-CR+CNN-SR~\cite{li2018learning}, CAR+EDSR~\cite{sun2020learned} and IRN~\cite{xiao2020invertible}.
\end{itemize}

To make the type III methods support non-integer scale factors $s$, we first bicubicly upscale $I_{\text{HR}}$ by $\frac{2}{s}$ or $\frac{4}{s}$ times for 1.0$<$$s$$<$2.0 and 2.0$<$$s$$<$4.0, respectively; and then feed it to the corresponding $\times$2/$\times$4 invertible downscaling models; finally bicubicly downscale the output image to the target resolution with scale factor $\frac{s}{2}$ or $\frac{s}{4}$.
Note that, this solution is not memory- and computation-efficient since it increases the resolution of images before feeding into the network and decreases the resolution after getting the network's output.
Differently, our AIDN can natively support arbitrary scale factors, without any pre- or post-processing.

\vspace{-2mm}
\paragraph{Quantitative results.}
Table~\ref{tab:SOTA_comparison} shows the PSNR of reconstructed HR images $\hat{I}_\text{HR}$ produced by multiple methods with various scale factors on five benchmark datasets.
Our AIDN significantly outperforms the type I and II methods, which confirms that jointly modeling the downscaling and upscaling process is beneficial.
For the 
type III methods
that jointly optimize the downscaling and upscaling, our AIDN still outperforms them by a large margin for non-integer scale factors on all the five benchmark datasets, while achieving comparable results for integer scale factors. 
Note that, results of the type III methods for different scale factors are produced by multiple trained models, \eg, IRN-$\times$2 and IRN-$\times$4, while results of our AIDN are generated by a universal model.
More importantly, as shown in Figure~\ref{fig:intro}, the performance of our AIDN changes much more smoothly across scale factors, compared with the type III methods.
This demonstrates the effectiveness of our CRM for invertible image downscaling with arbitrary scale factors.
Besides, the number of parameters of our AIDN is very small compared with other methods, which indicates the efficiency.
%

\vspace{-2.5mm}
\paragraph{Qualitative results.}
To further qualitatively evaluate the reconstructed HR images $\hat{I}_\text{HR}$ for non-integer scale factors, we compared our AIDN with the conventional Bicubic interpolation (BI+BI) and state-of-the-art scale-fixed invertible downscaling method, IRN~\cite{xiao2020invertible}, in Figure~\ref{fig:visual}.
Although the results of IRN were produced by multiple models, we can see results of our AIDN still have better perceptual quality and fewer artifacts, \eg, our method faithfully recovers both the letters and the flat region in the `ppt3' case, while BI+BI suffers from blurriness for the letters and IRN incurs color distortion in the flat region; and in the `img092' case, our method produces more accurate structure patterns than both BI+BI and IRN. 
The high fidelity of our results for various scale factors demonstrates the success of information embedding and the CRM.

\vspace{-2mm}
\subsection{Evaluation on Downscaled LR Images}
\begin{figure}[!t]
    \centering
    \includegraphics[width=0.93\linewidth]{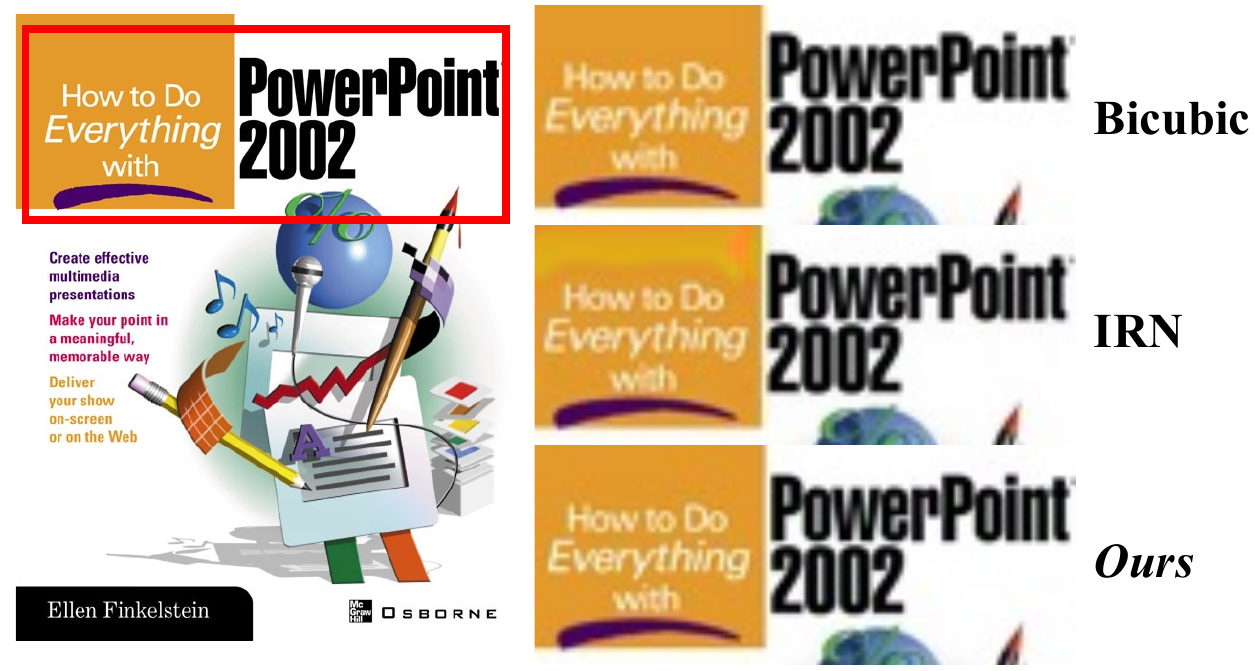}
    \vspace{-4mm}
    \caption{Visual comparisons of the downscaled image $\hat{I}_\text{LR}$ for the `ppt3' image in the Set14~\cite{zeyde2010single} with the scale factor of $\times$3.9.}
    \vspace{-6.5mm}
    \label{fig:LR_visual}

\end{figure}

Then, we evaluated the quality of downscaled LR images $\hat{I}_\text{LR}$, by measuring the SSIM between them and bicubic-downscaled images.
And we compared the results of IRN and our AIDN in Table~\ref{tab:evaluation_LR}, we can see both of them have extremely high SSIM values.
%
Besides, the qualitative comparison in Figure~\ref{fig:LR_visual} shows the perceptual quality of our results is even better than IRN, \ie, IRN incurs color distortion in the flat region while the result of our AIDN is free of the distortion.
It indicates our downscaled images are almost the same as the conventional bicubic-downscaled ones.

\begin{table}
  \caption{
  SSIM between the bicubic-downscaled images and the results produced by IRN and our AIDN on the B100~\cite{martin2001database} dataset.
}
\vspace{-3.5mm}
  \label{tab:evaluation_LR}
\centering
  \begin{tabular}{l|cccccc}
  \toprule[0.8pt]
    \multicolumn{1}{c|}{Method}&$\times$1.6    & $\times$2.1   & $\times$2.6 & $\times$3.1  & $\times$3.6   &$\times$3.9\\
  \hline
  IRN  & 0.9963 & 0.9949 & 0.9942 & 0.9936 & 0.9932 & 0.9930\\
  Ours & 0.9951 & 0.9937 & 0.9924 & 0.9915 & 0.9909 & 0.9907\\
  \bottomrule[0.8pt]
  \end{tabular}

\vspace{-4mm}
\end{table}

\begin{table*}[th]
  \caption{
  PSNR/SSIM results of the reconstructed HR images produced by the Bicubic interpolation, the variants of our method, and our full method on the Set5~\cite{bevilacqua2012low} dataset. The \best{best} and \seco{second-best} results are marked in bold and underline, respectively.
}
\vspace{-3mm}
  \label{tab:ablation_study}
  \centering
  \resizebox{\textwidth}{!}{
  \begin{tabular}{l|ccccccc}
  \toprule[0.8pt]
\multicolumn{1}{c|}{Method} & $\times$2 & $\times$2.3 & $\times$2.6 & $\times$3 & $\times$3.3 & $\times$3.6 &$\times$4           \\
  \hline
Bicubic  & 33.66/0.9299               & 32.43/0.9121               &  31.46/0.8944             & 30.39/0.8692        & 29.62/0.8491 & 29.03/0.8333 & 28.42/0.8104\\
AIDN$_\text{w/o CRM}$  & \best{45.73}/\best{0.9921} & 40.25/0.9766               &  39.33/0.9708             & \seco{38.62}/\best{0.9642} & 36.86/0.9515 & 36.20/0.9456 & 35.53/0.9381\\
AIDN$_\text{w/o content}$  & 44.06/0.9866               & \seco{41.81/0.9794}               &  \seco{40.25/0.9727}             & \seco{38.62}/0.9639        & \seco{37.68/0.9573} & \seco{36.85/0.9513} & \seco{35.93/0.9429}\\
AIDN (full method) & \seco{44.13/0.9868}               & \best{41.86}/\best{0.9797} & \best{40.31}/\best{0.9729} & \best{38.70}/\seco{0.9640} & \best{37.79}/\best{0.9579} & \best{36.98}/\best{0.9516} &\best{36.06}/\best{0.9436}\\
  \bottomrule[0.8pt]
  \end{tabular}}

\vspace{-3mm}
\end{table*}

\vspace{-2mm}
\subsection{Ablation Study}
To verify the effectiveness of some key designs in our AIDN, we conducted ablation experiments on the Set5~\cite{bevilacqua2012low} dataset by considering the following methods:
\begin{itemize}
\vspace{-1mm}

    \item \textbf{Bicubic}: down-/up-scaling with the Bicubic interpolation;

    \item \textbf{AIDN$_\text{w/o CRM}$}: the scale-fixed variant of our AIDN by removing the scale-aware feature adaption module in the feature extractor and replacing the CRM with the PixelShuffle;

    \item \textbf{AIDN$_\text{w/o content}$}: the variant of our AIDN that is not content-adaptive by making the CRM only conditioned on scale factors; and 

    \item \textbf{AIDN}: our full method.

 \vspace{-1mm}
\end{itemize}

As shown in Table~\ref{tab:ablation_study}, the PSNR/SSIM values of the Bicubic method are much lower than others, which confirms jointly modeling the downscaling and the upscaling can significantly improve the performance.
Comparing the results of {AIDN$_\text{w/o CRM}$} and {AIDN$_\text{w/o content}$}, we can see the {AIDN$_\text{w/o content}$} performs better for the non-integer scale factors while the {AIDN$_\text{w/o CRM}$} performs slightly better for the integer factors.
Note that, the {AIDN$_\text{w/o CRM}$} is designed for fixed integer scale factors, so we adopt a similar solution as the type III methods explained in Sec.~\ref{subsec:HR} to support non-integer scale factors, which means multiple trained models have to be stored in place.
Most importantly, our full method outperforms the {AIDN$_\text{w/o content}$} for both integer and non-integer scale factors.
It demonstrates applicability of our framework and the effectiveness of our content-adaptive design for the conditional resampling module (CRM).

\begin{figure}[!t]
    \centering
    \includegraphics[width=0.82\linewidth]{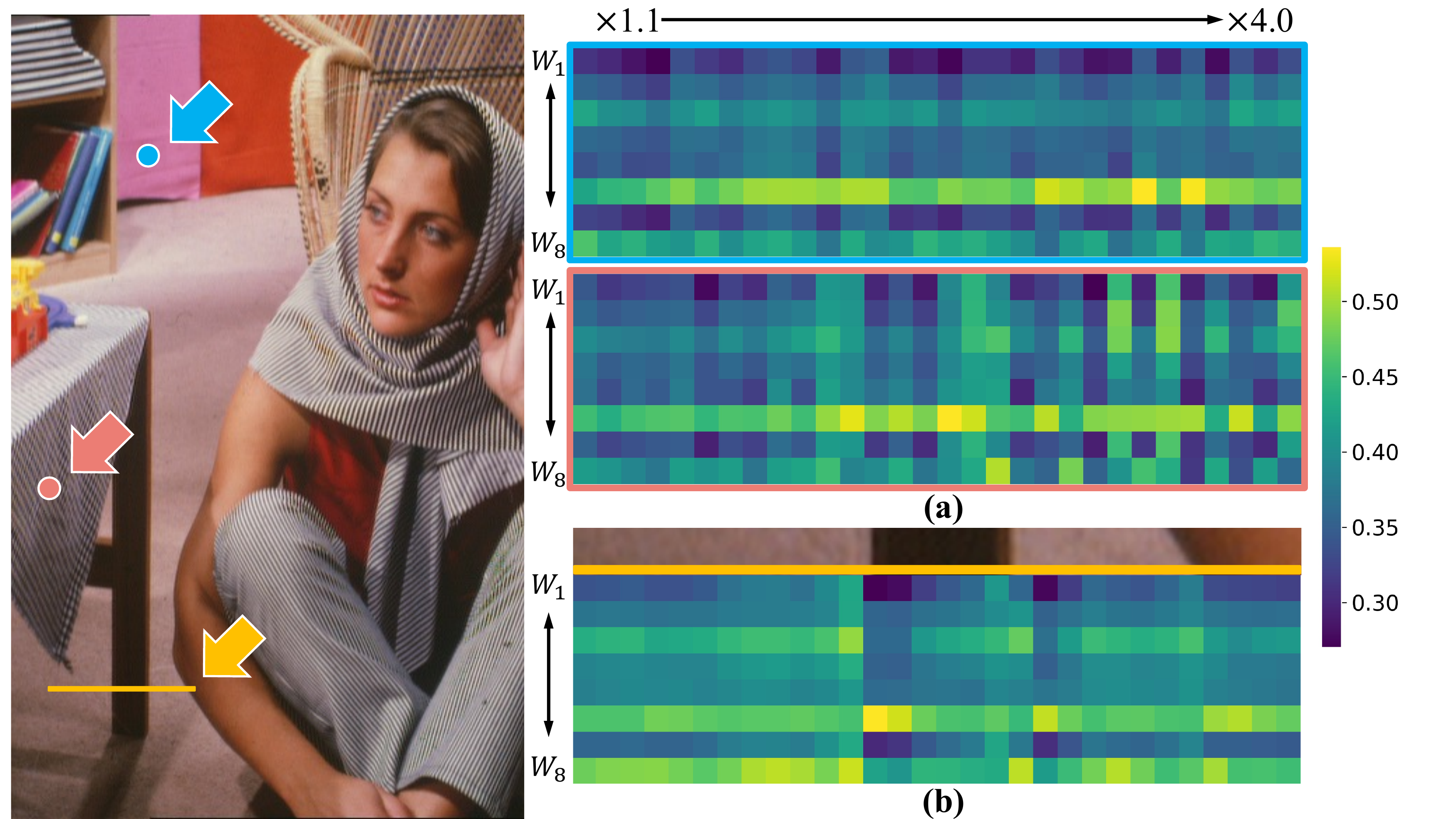}
\vspace{-2mm}
    \caption{Visualization of the routing weights produced by the CRM in the decoder with varying scale factors (a), and image contents (b). The blue and red dots indicate two sampling locations for fixing the image content while varying the scale factor. The yellow line is the sampling locations for fixing the scale factor to be $\times$4 while changing the image content.
     }
\vspace{-5mm}
    \label{fig:visual_routing}
\end{figure}

\vspace{-2mm}
\subsection{Discussion}

\paragraph{Visualization of routing weights.}
\label{sec:routing}
To verify the effectiveness of the CRM for dynamically producing resampling kernels for different scale factors and image contents, we visualized the routing weights produced by the CRM in the decoder network, with various scale factors and image contents.
We first fixed the image content and changed the scale factors to see the resulting routing weights in the decoder.
As shown in Figure~\ref{fig:visual_routing} (a), for the same location on the image, eight experts are activated differently when the scale factor varies, and the tendency to change is not the same for different locations \eg, the two marked locations with blue and red dots.
Then, we fixed the scale factor to be $\times$4 and uniformly sample the locations on a line in the image (Figure~\ref{fig:visual_routing} (b)) to observe the resulting routing weights.
We can see the eight routing weights are almost the same within the flat region, while changing sharply along with the edge in the image.
It demonstrates our CRM can produce adaptive resampling kernels according to both the scale factor and the image content.

%

\begin{figure}[!t]
    \centering
    \includegraphics[width=1\linewidth]{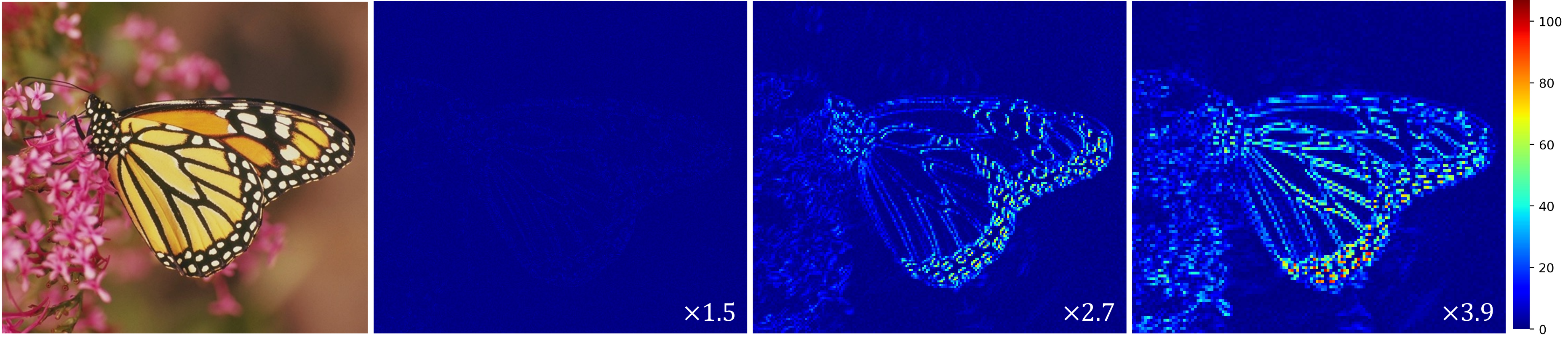}
\vspace{-6mm}
    \caption{Difference maps between our embedded LR images and bicubic-downscaled ones under various scale factors. We show the original HR image on the left for reference.
     }
\vspace{-4mm}
    \label{fig:difference_map}
\end{figure}


\vspace{-2mm}
\paragraph{HR information embedding.}
To further explore the embedding mechanism of HR information in the LR image, we visualized the difference maps between our embedded LR images and bicubic-downscaled ones under various scale factors in Figure~\ref{fig:difference_map}. 
For convenience, we resized difference maps into the same resolution.
We can observe that the differences are trivial when the scale factor is near to one, while the differences are concentrated on the edges when the scale factor is faraway from one, and the differences become more significant with the scale factors increasing.
This is because there is very little HR information need to be embedded when the scale factor is near to one, while more HR information is demanded to be embedded when the scale factor increases. 
%

\vspace{-2mm}
\paragraph{Limitation.}
Since our AIDN encodes the HR information into the downscaled LR image, the quality of the restored HR image highly depends on the embedded information. 
Therefore, the downscaled LR image by our AIDN may not be resistant to general image manipulations, such as JPEG compression and editing. 
But many studies~\cite{wu2021embedding,xing2021invertible,du2021invertible,liu2021jpeg,hu2020mononizing} have shown that incorporating the compression during training can significantly boost the robustness to compression, which is the potential direction of our future work. 
%

\vspace{-2mm}
\section{Conclusion}
We present a scale-arbitrary invertible image downscaling network (AIDN) to natively downscale high-resolution (HR) images with arbitrary scale factors, such that the downscaled images can meet the resolution restriction of conventional social media platforms. 
%
Meanwhile, the HR information is embedded in the downscaled low-resolution (LR) counterparts; thus, our AIDN can also restore the original HR images with high-quality solely from the LR images whenever users want to explore the details of the distributed images.
Our technical contribution is the proposed Conditional Resampling Module (CRM) that can dynamically resample feature maps to the target resolution according to both the scale factor and the image content.
Extensive ablation experiments and the visualization of the routing weights verified our design intent for the CRM.
Both quantitative and qualitative results demonstrate our AIDN achieves top performance for invertible image downscaling with
both arbitrary integer and non-integer scale factors.

\bibliographystyle{ACM-Reference-Format}
\bibliography{paper}
\end{document}